\documentclass{bmvc2k}

\usepackage{caption}
\title{Weak Supervision for Label Efficient Visual Bug Detection}

\addauthor{Farrukh Rahman}{farrukh.rahman@microsoft.com}{1}

\addinstitution{
Xbox Studios QuAIL
}

\usepackage{graphicx}
\usepackage{tabularx}
\usepackage{booktabs}

\usepackage{colortbl}

\definecolor{Gray}{gray}{0.9}

\runninghead{Rahman}{Weak Supervision for Label Efficient Visual Bug Detection}

\begin{document}

\maketitle

\begin{abstract}
 As video games evolve into expansive, detailed worlds, visual quality becomes essential, yet increasingly challenging. Traditional testing methods, limited by resources, face difficulties in addressing the plethora of potential bugs. Machine learning offers scalable solutions; however, heavy reliance on large labeled datasets remains a constraint. Addressing this challenge, we propose a novel method, utilizing unlabeled gameplay and domain-specific augmentations to generate datasets \& self-supervised objectives used during pre-training or multi-task settings for downstream visual bug detection. Our methodology uses weak-supervision to scale datasets for the crafted objectives and facilitates both autonomous and interactive weak-supervision, incorporating unsupervised clustering and/or an interactive approach based on text and geometric prompts. We demonstrate on first-person player clipping/collision bugs (FPPC) within the expansive Giantmap game world, that our approach is very effective, improving over a strong supervised baseline in a practical, very low-prevalence, low data regime (0.336 $\rightarrow$ 0.550 F1 score). With just 5 labeled "good" exemplars (i.e., 0 bugs), our self-supervised objective alone captures enough signal to outperform the low-labeled supervised settings. Building on large-pretrained vision models, our approach is adaptable across various visual bugs. Our results suggest applicability in curating datasets for broader image and video tasks within video games beyond visual bugs.
\end{abstract}

\vspace{-0.5cm}
\section{Background \& Introduction}
\label{sec:intro}
\begin{figure}
  \centering
  \scalebox{0.9}{%
  \includegraphics[width=\linewidth]{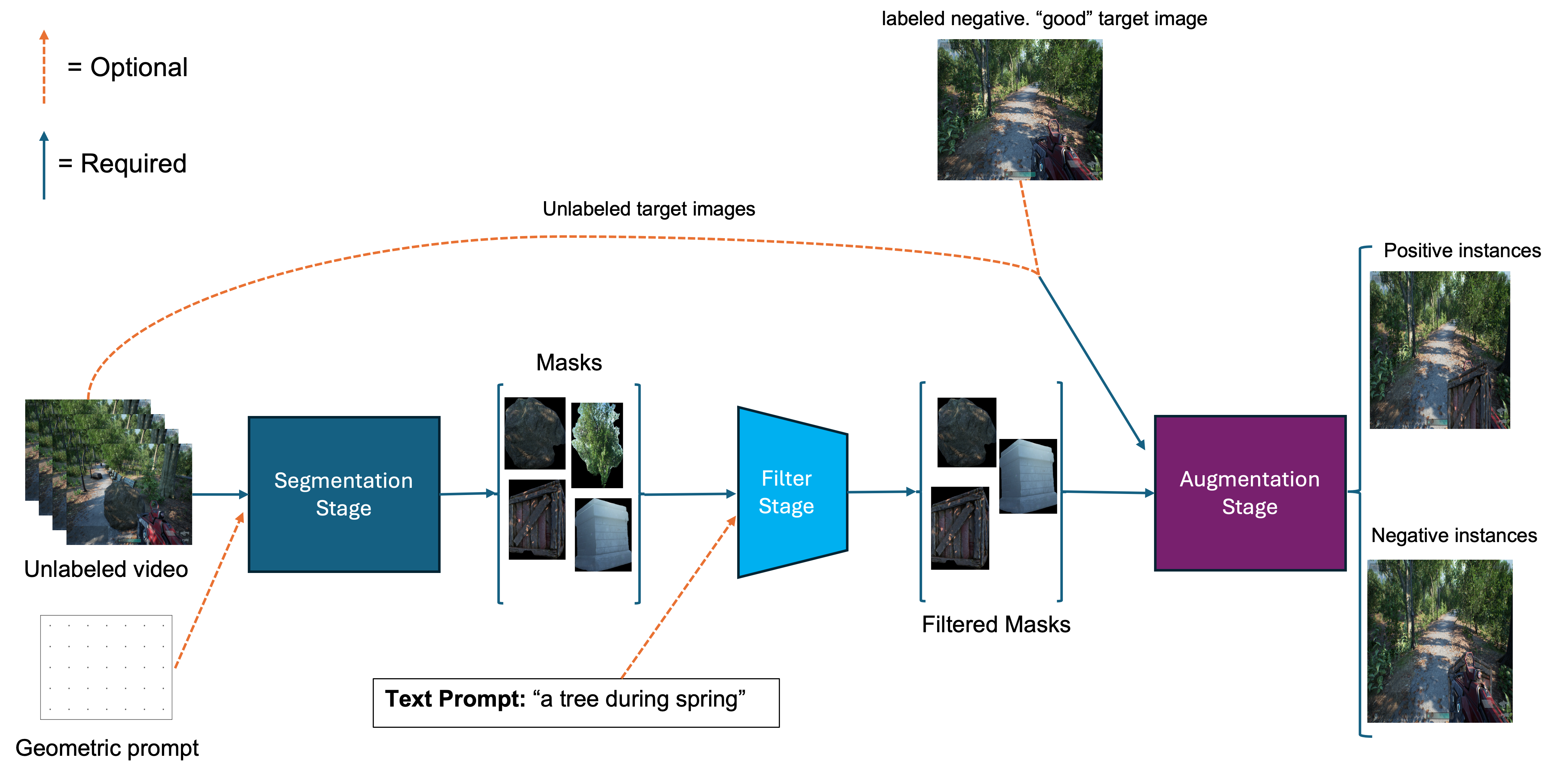}
  }
  \caption{
  General overview of our method: \textbf{1. Segmentation Stage:} Given unlabeled gameplay video, we apply a geometric promptable segmentation model (SAM) to automatically extract masks. \textbf{2. Filtering Stage:} The obtained masks are then filtered either in an unsupervised manner and/or optionally via text-interactive filtering using text-image model (CLIP). \textbf{3. Augmentation Stage:} Labeled \textit{`good'} target instances, and/or unlabeled target instances, are augmented using the filtered masks producing samples used to train a surrogate objective.}
  \label{fig:method1}
\end{figure}

\vspace{-0.2cm}
 Visual quality in video games is one of the key drivers of satisfaction with customers. With modern games transitioning towards expansive, open worlds with intricate visuals and systems, the potential for bugs rapidly grows. Traditional manual testing methods, constrained by time and resources, grapple with these challenges. Advances in Computer Vision (CV) and Machine Learning (ML) present promising alternatives, offering automated and scalable visual testing solutions, thereby reallocating resources to explore other game dimensions \cite{nantes2008framework}. Notably, the success of deep learning in CV is largely credited to extensive labeled datasets \cite{deng2009imagenet, lin2014microsoft}, often curated from the vast quantitites of digital content on the web. However, curating these massive labeled datasets for a single game is impractical. Manual capturing and labeling of visual bugs at scale would render detection methods redundant, more so given the rarity of such bugs. Computer vision based methods recently proposed facilitate automated visual testing at scale by \textbf{1.} leveraging game engines to increase data availability amenable to deep learning approaches \cite{wilkins2022world, lod_cnn_matilda, ling2020usingcnn, taesiri2020video} and/or \textbf{2.} using anomaly detection based approaches treating bugs as out of distribution (OOD) occurrences from normal frames \cite{wilkins2020metric}. While access to game engines endow greater data availability and control over diversity,  the non-stationary nature of games requires an evolving set of data generated for every new asset across multiple factors (environment, lighting, etc.) for any given game title. Additionally, the limited testing window in a game development cycle places emphasis on the speed of adaptation of any particular detection method. Addressing these challenges, we propose using unlabeled gameplay video paired with domain-specific augmentation techniques to derive objectives for visual bug detection models. This strategy is useful in the low-labeled settings often present during game development. Specifically, our method (fig.~\ref{fig:method1}) utilizes large-pretrained vision models \cite{radford2021learningCLIP, kirillov2023segment} also termed foundation models \cite{bommasani2021foundation} along with domain specific augmentation strategies motivated by \cite{ghiasi2021cutpaste} to formulate self-supervised objectives for which we scale datasets through weak-supervision.
 \textit{Self-supervised learning} (SSL) seeks to learn from unlabeled data through optimization of a defined surrogate objective, which is then transferred to downstream target tasks \cite{balestriero2023cookbook}. SSL has shown to learn transferable representations across multiple domains including CV \cite{chen2020simple, gidaris2018unsupervisedrotnet, he2020momentum, he2021masked}. \textit{Weak Supervision} leverages noisy annotation sources to expediently generate and scale noisily labeled datasets \cite{ratner2017snorkel, zhang2022surveyweak}, recently demonstrating effectiveness in training large-scale models across multiple domains \cite{radford2022whisper,radford2021learningCLIP, kirillov2023segment}. \textit{Interactive weak supervision} furthers this via an interactive process \cite{boecking2021interactive, ouyang2022traininghuman} merging domain expertise with scalability of weak supervision. Our methodology uses domain-specific SSL objectives that are scaled through weak supervision, leveraging large pre-trained models and integrating text and geometric prompts for efficient interaction. We demonstrate the generality of our method by targeting multiple visual bug-types, egocentric/first-person player clipping and texture issues. Moreover, from analyzing our results we suggest our method can be adapted to curate extensive datasets for a range of image and video analysis tasks in video games, extending beyond visual bug detection. 

\vspace{0.15cm}
The main contributions of this work are summarized as follows:

\begin{enumerate}
    \item \textbf{Empirical Observations on ViT Performance:} We observe that when trained with a self-supervised method, DINOv1, \textbf{a.} ViT surpasses traditional ResNet architectures, and \textbf{b.} DINO rivals the performance of supervised pre-training on IN1K in low-labeled and few-shot settings for visual bugs.
    \vspace{-0.2cm}
    \item \textbf{Development of a Novel Methodology:} Building on the aforementioned observations, we introduce a flexible  technique that uses weak-supervision to scale a self-supervised objective. This approach melds zero-shot segmentation (Segment-Anything) and domain-specific augmentations. Notably, our method consistently delivers strong results across practical out-of-distribution (OOD) contexts.
    \vspace{-0.2cm}
    \item \textbf{Extension via Clustering and Filtering:} We integrate a filtering step to enhance performance using unsupervised clustering and text-image models such as CLIP, offering two distinct avenues: \textit{automated}  or \textit{text-interactive} weak supervision. The latter enables non-ML practitioners to add preferable inductive biases to guide the system through simple text and geometric prompts.
    \vspace{-0.2cm}
    \item \textbf{Efficient Dataset Curation:} Our research underscores the potential for efficient dataset curation. Given a handful of labeled "good" exemplars and a small amount of domain expertise, datasets can be curated autonomously. From these we can craft objectives for standalone few-shot models, pre-training, or multi-task scenarios in low-data regimes.
  
\end{enumerate}

\vspace{-0.75cm}
\section{Approach}
 Several practical challenges arise in the domain of visual bug detection, which shape our objectives. Firstly, there is the issue of limited labeled data. The timeframes during which visual testing can be conducted are narrow, especially with fresh content. Methods amenable to low-data regimes and/or faster transfer learning are highly coveted. A second is access to source code; engines such as \cite{unity3d, unrealengine} continue to integrate ML features increasing data for models to consume, yet this is impractical to scale across every game (eg. building hooks into every new sub-release of a given game).  We seek methods that can be applied in scenarios where access to the source code is not guaranteed. Related to this is the notion of out-of-distribution (OOD) scenarios, namely that even if we could gather data at a given point during development, as new content is added we want our model to adapt to new scenarios with minimal new data. An additional point here is that our input data during test time is constrained to RGB frames. Moreover, a third practical constraint is the notion that bugs are often rare and performant methods in low-prevalence scenarios are valuable.

\subsection{Datasets}
\begin{figure}
    \centering
     \scalebox{0.5}{%
    \begin{minipage}{.39\linewidth}
        \centering
        \includegraphics[width=\linewidth]{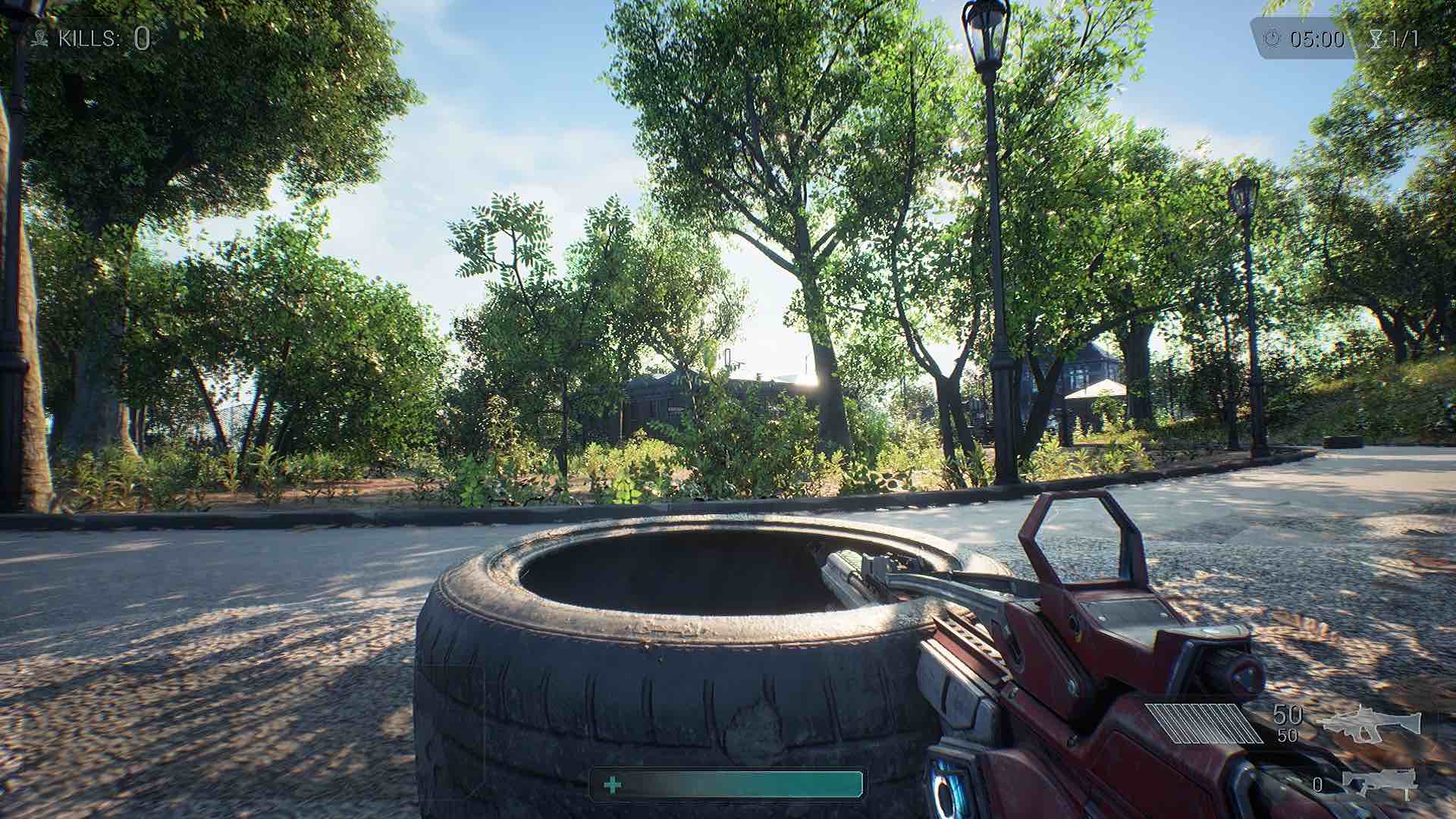}
    \end{minipage}
    \hfill
    \begin{minipage}{.39\linewidth}
        \centering
        \includegraphics[width=\linewidth]{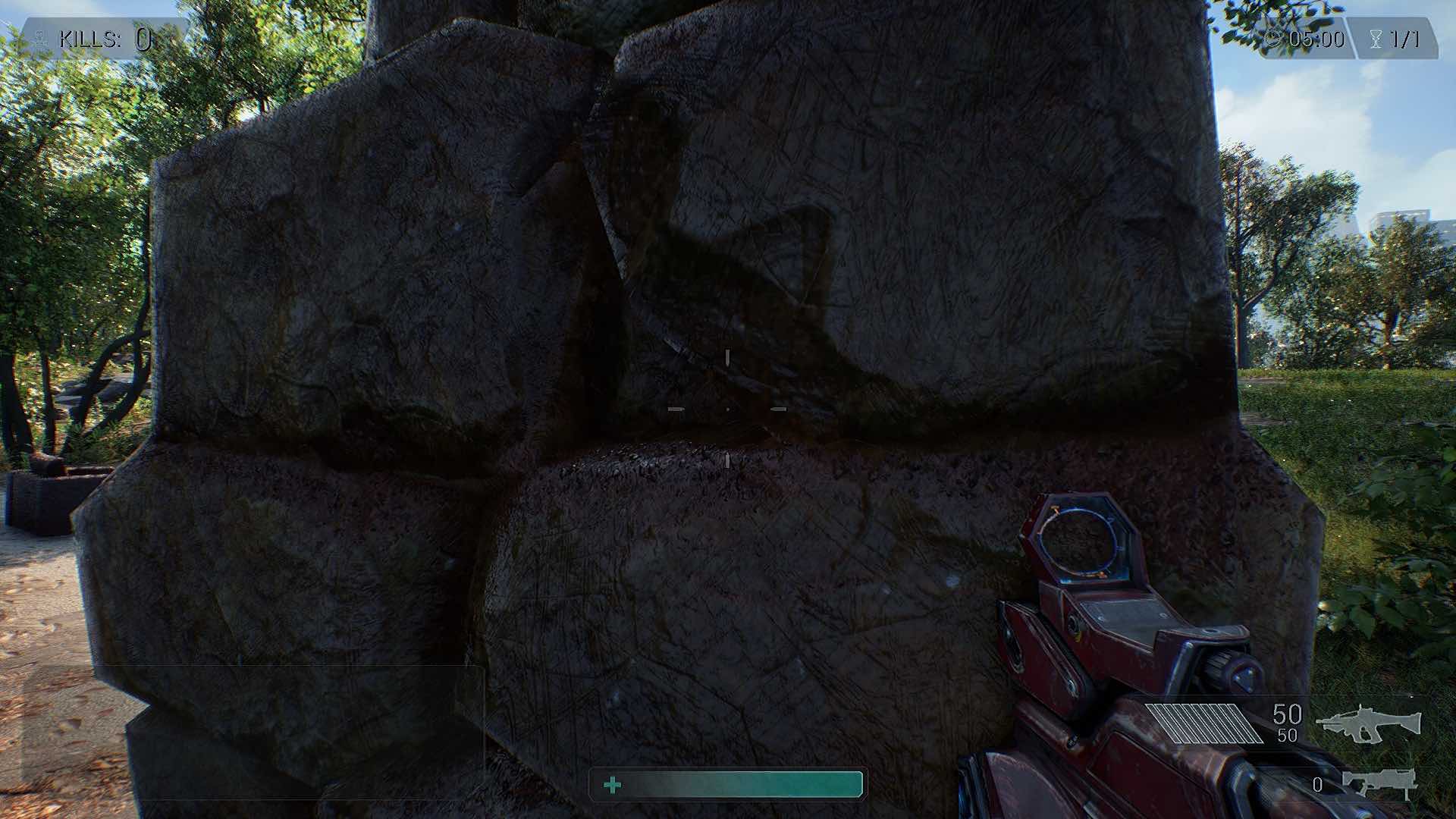}
    \end{minipage}

    \begin{minipage}{.39\linewidth}
        \centering
        \includegraphics[width=\linewidth]{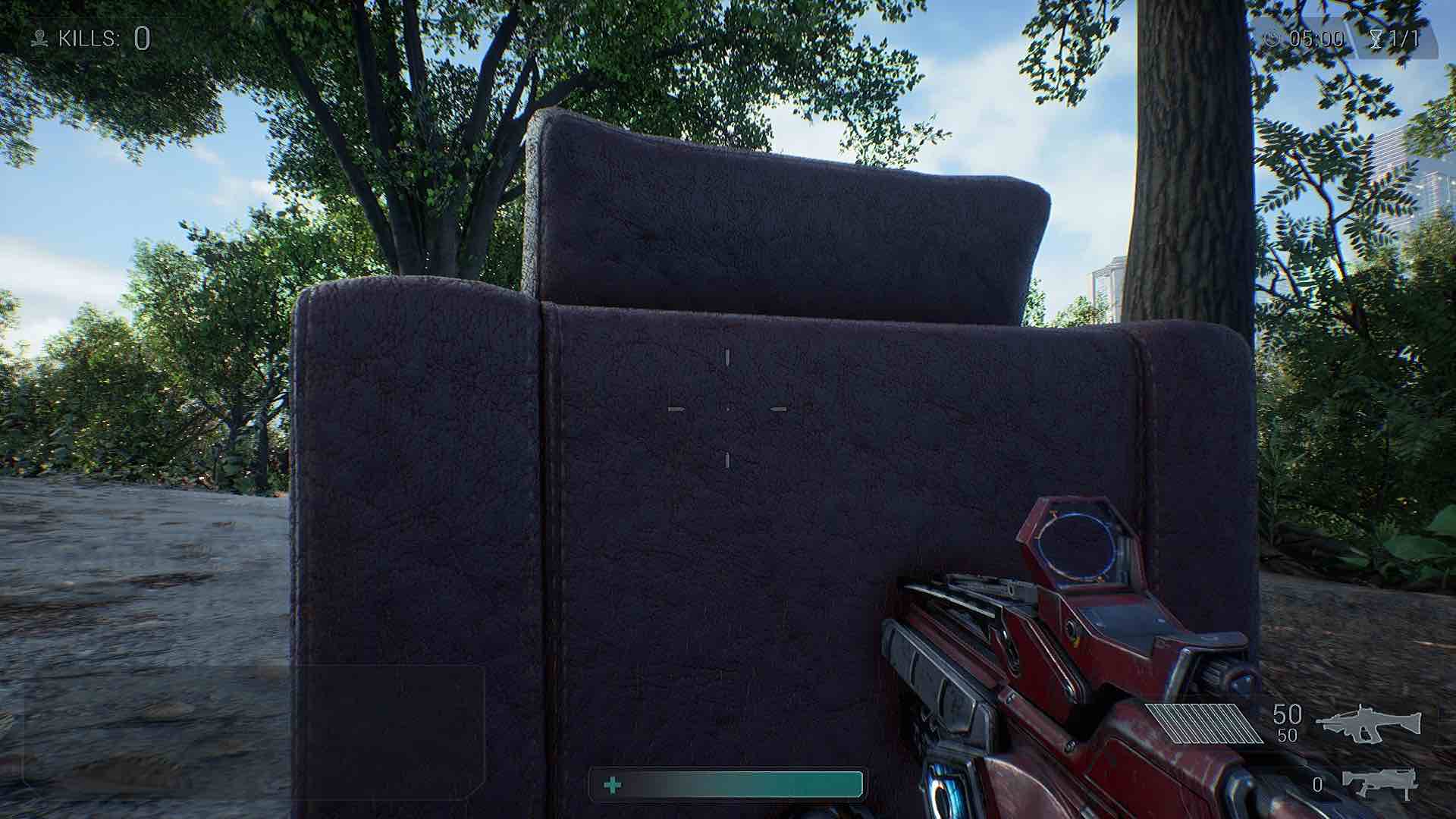}
    \end{minipage}
    \hfill
    \begin{minipage}{.39\linewidth}
        \centering
        \includegraphics[width=\linewidth]{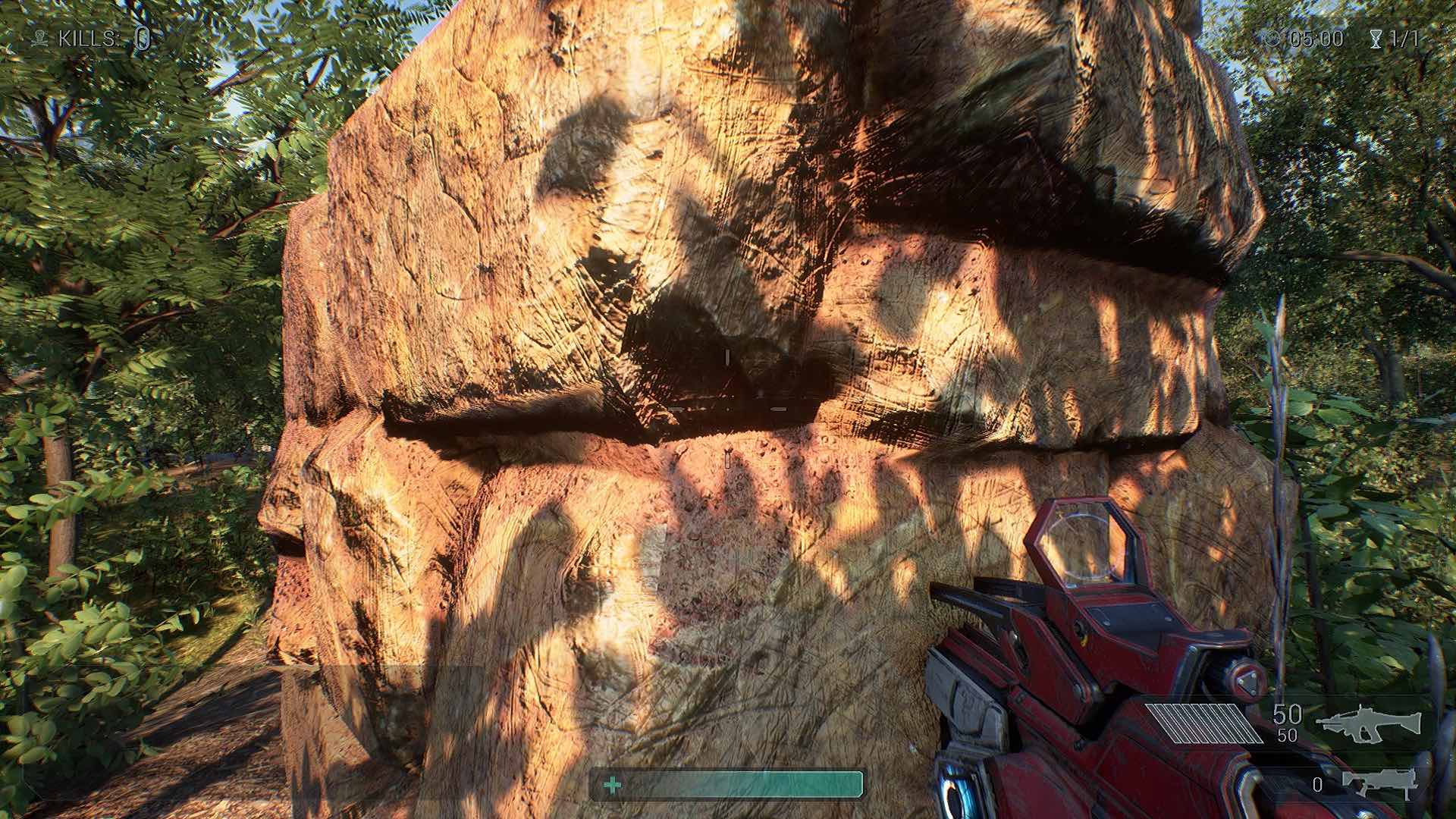}
    \end{minipage}

    \hfill
    \begin{minipage}{0.4\linewidth}
        \centering
        \includegraphics[width=1.05\linewidth]{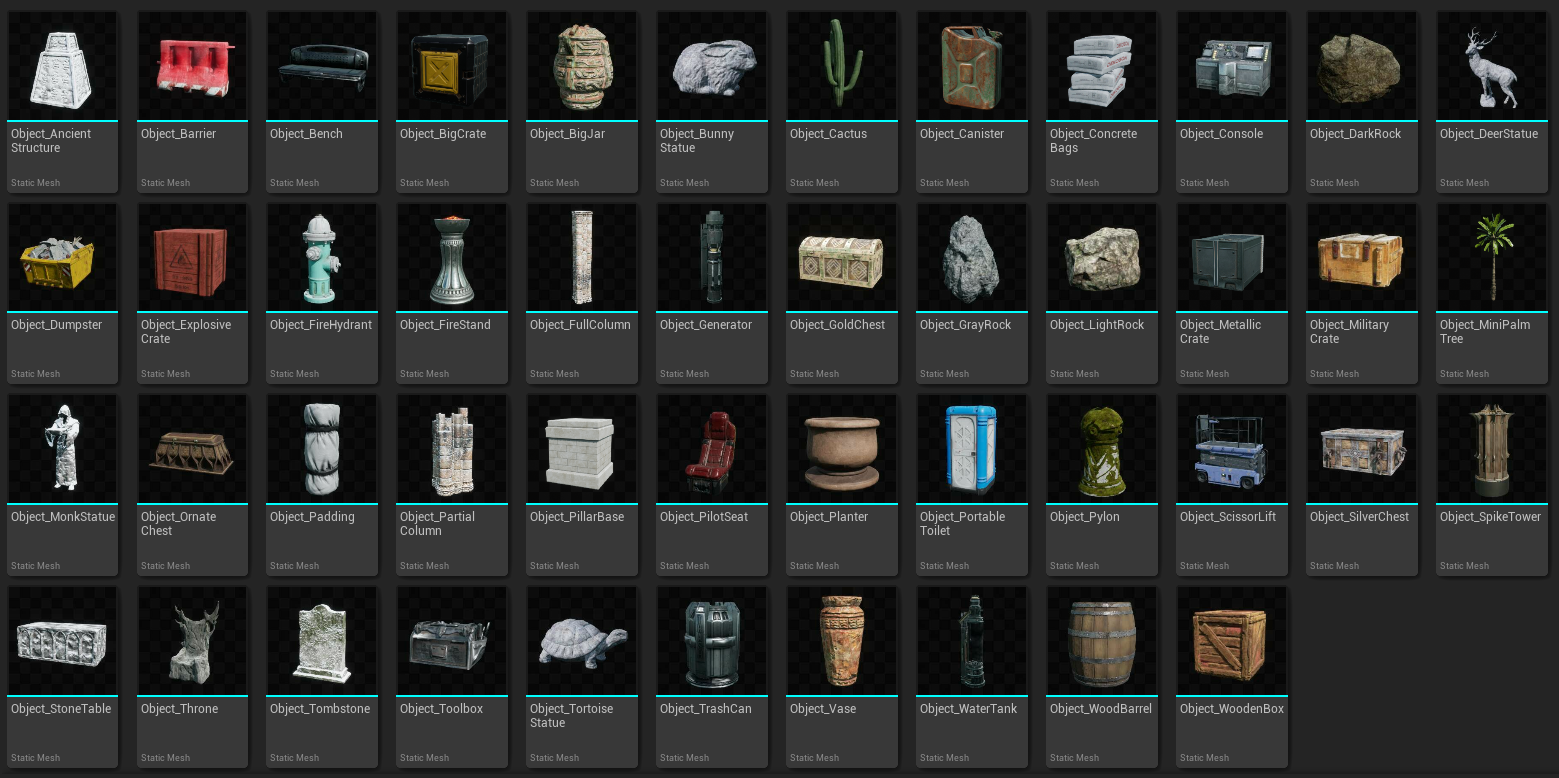}
        \label{fig:gm50}
    \end{minipage}
    }
    \caption{(left 4 images) in-distribution Clipping examples from GM-4 set. (Right image) 46 Out-of-distribution objects added in GM-50. }
    \label{fig:clipExamplesGM4}
\end{figure}

 We use the Giantmap-5 (now GM4 as one object was removed) environment and active area as introduced in \cite{abdelfattah2023preferenceconditioned}, developed in Unreal \cite{unrealengine}. We further extend it by introducing 46 new objects of interest (OOI) shown in fig.~\ref{fig:clipExamplesGM4}. In this study, we treat the Giantmap environment as our target video game title for our chosen visual bug, first-person player clipping (FPPC). FPPC manifests when collision meshes for either the player or object are set incorrectly or naively creating visual aberrations that would not occur in the physical world, see fig.~\ref{fig:clipExamplesGM4} of FPPC on the 4 objects on our GM4 environment. From this environment we create  i.i.d. screenshots programmatically by first generating an object distribution over the map with a specified density, then spawning the player near objects within a certain distance from the center of the object to sample varied clipping and normal samples. This capability allows us to scale data generation significantly however we seek to push the boundaries of label efficiency treating Giantmap as our target title. How far can we push in-distribution performance and how does it fare in OOD scenarios? To this effect, we constrain training data to 15 total samples for GM4-tiny dataset and 156 samples for GM4-base dataset whilst generating 3k validation and test in-distribution sets. Moreover, we generate a low-prevalence (0.007) video \textit{deployment} set on GM50 (4 ID + 46 OOD objects) to evaluate our methods, in effort to mimic what a developer might collect from automated or human play testing.  Additionally we gather separate human gameplay on GM50 to use with the small amount of labeled data generated. In summary, we are given a small amount of i.i.d screenshot in-distribution data, Unlabeled OOD video, and are expected to evaluate on an OOD, low-prevalence video. \label{datasets}

\vspace{-0.2cm}
\subsection{Method}
\label{weaklysupervisedMethod}
   Our method can be viewed as a self-supervised objective scaled through weak-supervision. As shown generally in fig.\ref{fig:method1}, it consists of 3 main stages, and is described in more detail below. We use the first-person player clipping task to show the efficacy of the approach as it is a challenging visual bug.

   \textbf{Segmentation Stage:} Given unlabeled gameplay video of a target video game, we apply a pre-trained, promptable segmentation model SAM \cite{kirillov2023segment} to extract masks in an automated manner. SAM takes as input an image and one or more geometric prompts. In absence of any prompt, points are placed uniformly across the image which represents the automatic/zero-shot segmentation prompt. Priors can be injected into the prompt to guide SAM to ignore or further sample certain regions of the input frame.

   \textbf{Filtering Stage:}   \label{filtering} Since the environment is an outdoor park set in the spring, certain semantic visual features are abundant, eg. trees, walking trails, or grass. We develop a filtering \& deduplication step using CLIP \cite{radford2021learningCLIP}, a text-image model to extract embeddings of each masked region. For \textit{autonomous filtering}, we first cluster embeddings using Hierarchical Agglomerative Clustering (HAC) \cite{Johnson1967HierarchicalCS, scikit-learn}, then re-sample masks from each cluster aiming to balance the mask distribution. For \textit{interactive filtering}, a user may apply prior knowledge to select for or against certain masks via a text prompt, after which we perform clustering. The text prompts are embedded using the CLIP text encoder and cosine distances are computed with each mask embedding. Text-prompting capability can autonomously incorporate prior knowledge; for instance, if prior knowledge indicates that foliage, trees, and grass aren't relevant, text-prompts around these semantics can be cached and applied as pre-processing prior to unsupervised clustering. The final set of masks represents the set of semantics on the playthrough/game-level expected to be observed in a scene, intrinsically making them good candidates for visual bug augmentation. Moreover, the policy under which the data is collected also contributes to the mask distribution; we make an explicit assumption that the semantics of a target game are captured in the unsupervised playthroughs. 
  

\textbf{Augmentation Stage:} Masks along with target images are used to create a self-supervised objective through domain-specific augmentation. Target images can be obtained from a small labeled set, or directly from the source unlabeled data. As the masks represent semantics of the target game we utilize them to create augmented positive examples denoting bugs and negative examples denoting \textit{"normal"} or \textit{"no-bug"}. If variants of a particular bug exist (e.g., stretched vs low-res texture), multiple classes can be augmented. As the method is tailored to the downstream task, in certain scenarios, the source and target image can be identical. Our method is flexible and can be applied across a variety of visual bug types. 

\begin{figure}
\centering
\scalebox{0.9}{%
  \includegraphics[width=\linewidth]{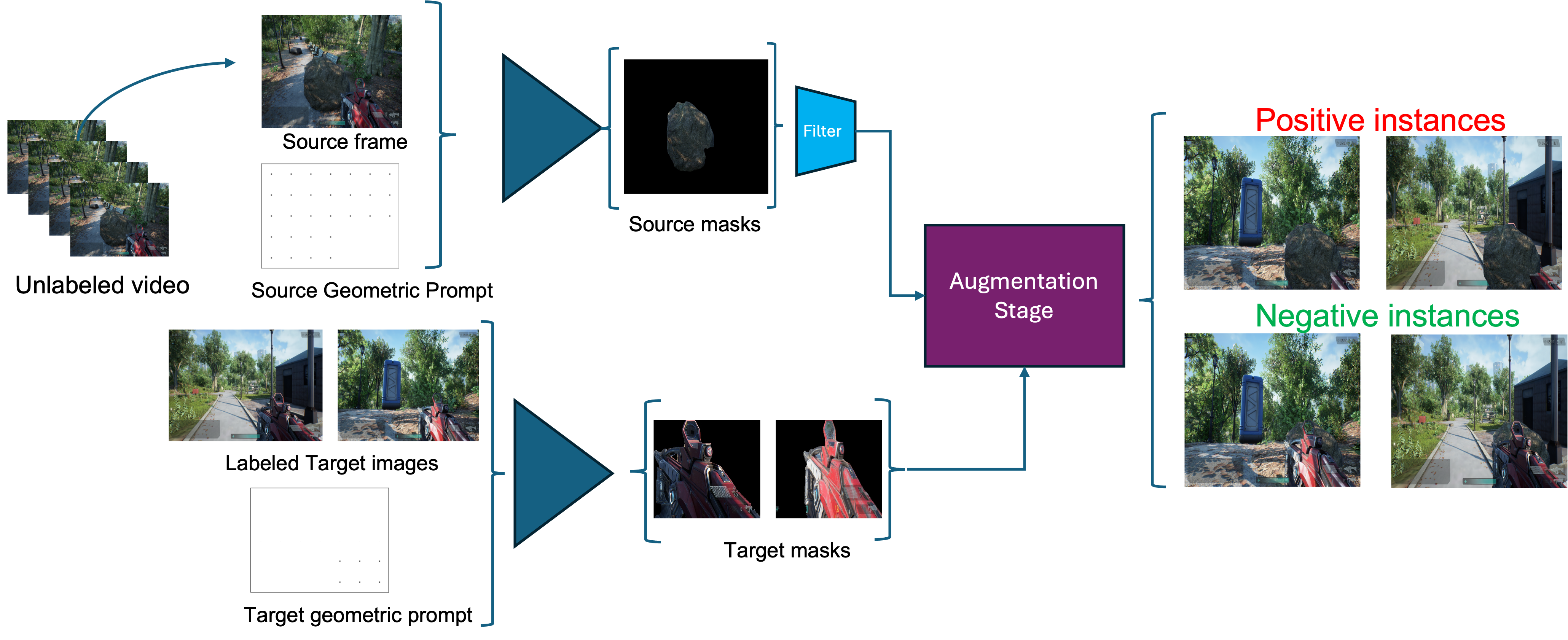}
}
  \caption{
  Our method from fig.\ref{fig:method1} instantiated w.r.t. first-person player clipping. From an unlabeled video, 5 target frames (2 shown) are labeled and processed by SAM (in dark blue) with geometric prompts. Source geometric prompts guide SAM to disregard the 'prior region' (i.e. weapon region), while target prompts emphasize only that region. After filtering, source masks, along with with target masks and target images, proceed to the augmentation phase. Here, positives are created by overlaying the source mask \textit{over} the target image's weapon area, while negatives are positioned \textit{behind} the weapon, respecting the target weapon mask. Classifying positives vs negatives serve as our self-supervised objective for FPPC.}

  \label{fig:methodClip}
  
\end{figure}

\textbf{First-person Weapon Clipping approach:} 
 We instantiate our general method for First-Person (or egocentric) player clipping (FPPC), fig.~\ref{fig:methodClip}. During segmentation prompting we prefer to ignore the bottom-right corner of the image typically where the weapon is placed; thus preventing saturating detected masks with weapon masks. From the unsupervised gameplay video, first the video was down-sampled temporally as videos naturally have visual information redundancy among adjacent frames. Semantic redundancy however is useful as the same object viewed from a different view increases both the probability of acquiring a good mask, as well as instance diversity.  \label{maskDatasets}From said subsample, two further sets are sampled, 300 frames to build a tiny dataset of 217 masks, and 20k frames to build a larger set of 17k masks. The filtering step is unchanged from fig.~\ref{fig:method1}.
 For our specific setting, we elect to paste the mask \textit{over} the weapon in a given target image. This creates a \textit{"peudo-clipping"}, or \textit{"weapon obstruction"} signal which we hypothesize is correlated with our target downstream clipping task. Conversely, the mask is copied \textit{under} the weapon (respecting the weapon's mask) to create a negative sample. In order to achieve this, we require labeled-\textit{good} images as targets.  We label 5 random frames from the human gameplay video and use them as target images\label{fewgoodshots}. Each target image is paired with each mask for 2 rounds of augmentation (pseudo-clip vs no-clip). During the augmentation, the source mask can be further augmented before it is pasted onto the target images. We apply random rotation and random horizontal flip augmentations. Post augmentation the tiny mask set generates 2.2k total samples, while large generates 170k; which are used to pre-train, multi-task and few-shot on our target task.

\vspace{-0.2cm}
\section{Experiments \& Results}

 \textbf{In-Distribution performance on GiantMap-4:} We report the in-distribution balanced test accuracy of the various architectures evaluated in tab.~\ref{tab:GM5Tiny},~\ref{tab:GM5Base}. We evaluate ResNet \cite{he2015deepres} variants and Vision Transformer (ViT) \cite{dosovitskiy2021image}. Within each architecture we further evaluate various pre-training methodology including supervised, weakly-supervised and self-supervised learning methods. Specifically, IN1k \cite{deng2009imagenet} supervised pre-training using the traditional \cite{he2015deepres} and A1 ResNet training recipe from \cite{wightman2021resnet, rw2019timm}, DINOv1 \cite{caron2021emerging} self-supervised pretext task (for both ResNet and ViT) as well as weakly-supervised CLIP's \cite{radford2021learningCLIP} ViT based image-encoder. We use a few-shot fine tuning approach given recent results indicating its superiority when training in these regimes \cite{tian2020rethinking, chen2020closer}. Moreover, we evaluate using a crop prior compared with the full frame. Specifically regarding FPPC, given it mainly manifests with the weapon, we can ignore the other parts of the frame. Naturally, the prior is significantly more data efficient, see tab.~\ref{tab:GM5Tiny}. In parallel, treating the problem as an object detection problem was also explored however the crop prior approach shows greater data efficiency given no regression of bbox coordinates is required (ref. supplemental). Our results show \textbf{1.} few-shot fine tuning can be efficient and \textbf{2.} when pre-trained, Vision transformers seem to outperform traditional CNNs in low-labeled settings, similar to observations in other visual domains \cite{naseer2021intriguing, rahman2022surprising, zhou2021convnets}. Moreover, we observe that self-supervised pretraining (DINOv1) is competitive or slightly surpasses supervised pretraining when transfer learning to our task. i.e., DINO is able to extract relevant features that transfer well into the low-data regime, tab.~\ref{tab:GM5Base}. Given our strong baseline for balanced low-labeled in-distribution performance, we select ViT pretrained on DINO as our backbone for all future experiments where we will evaluate in a challenging out of distribution (OOD), low-prevalence setting observed in practice. In this imbalanced setting, we use F1 score (harmonic mean of precision and recall) as our primary metric.  
\label{experiments}

\begin{table}
    \centering
    \begin{minipage}{0.48\linewidth}
        \centering
        \scalebox{0.5}{ 
        \begin{tabularx}{1.6\linewidth}{l l l r}
            \toprule
            \textbf{Model Architecture} & \textbf{Pretrain Method} & \textbf{Prior} & \textbf{Accuracy} \\
            \midrule
            \rowcolor{Gray}
            ResNet-50 & In1k sup & Crop & 0.811 $\pm$ 0.06 \\
            \rowcolor{Gray}
            ResNet-50 & In1k sup A1 & Crop & 0.796 $\pm$ 0.03\\
            \rowcolor{Gray}
            ResNet-18 & In1k sup & Crop & 0.753 $\pm$ 0.04\\
            \rowcolor{Gray}
            vit-base-16 & In1k sup & Crop & 0.913 $\pm$ 0.03 \\
            \rowcolor{Gray}
            vit-base-16 & CLIP & Crop & 0.949 $\pm$ 0.03 \\
            \rowcolor{Gray}
            vit-base-16 & DINO v1 & Crop & \textbf{0.952} $\pm$ 0.03 \\
            \rowcolor{Gray}
            vit-base-16 & In1k sup & Crop & 0.825 $\pm$ 0.03\\
            \midrule
            ResNet-50 & In1k sup & - & 0.733 $\pm$ 0.05 \\
            vit-base-16 & DINOv1 & - & 0.824 $\pm$ 0.05 \\
            vit-base-16 & CLIP & - & 0.675 $\pm$ 0.02 \\
            vit-base-16 & In1k sup & - & 0.738 $\pm$ 0.02 \\
            \bottomrule
        \end{tabularx}
        }
        \caption{In-distribution test performance for training on GM4-Tiny Dataset (15 total samples). Results over 3 trials.}
          \label{tab:GM5Tiny}
    \end{minipage}
    \hfill 
    \begin{minipage}{0.48\linewidth}
        \centering
        \scalebox{0.5}{ 
        \begin{tabularx}{1.6\linewidth}{l l l r}
            \toprule
            \textbf{Model Architecture} & \textbf{Pretrain Method} & \textbf{Prior} & \textbf{Accuracy} \\
            \midrule
            \rowcolor{Gray}
            ResNet-50 & In1k sup & Crop & 0.958 \\
            \rowcolor{Gray}
            vit-base-16 & In1k sup & Crop & 0.9657 \\
            \rowcolor{Gray}
            vit-base-16 & CLIP & Crop & \textbf{0.979} \\
            \rowcolor{Gray}
            vit-base-16 & DINOv1 & Crop & 0.976 \\
            \rowcolor{Gray}
            vit-base-16 & In1k sup & Crop & 0.9148 \\
            \rowcolor{Gray}
            ResNet-50 & DINOv1 & Crop & 0.9664 \\
            \midrule
            ResNet-50 & In1k sup & - & 0.922 \\
            vit-base-16 & DINOv1 & - & 0.967 \\
            vit-base-16 & CLIP & - & 0.89 \\
            vit-base-16 & In1k sup & - & 0.961 \\
            \bottomrule
        \end{tabularx}
        }
        \caption{In-distribution test performance for training on GM5-base Dataset, 156 total samples. $\pm$0.02 over 3 trials.}
          \label{tab:GM5Base}
    \end{minipage}
\end{table}

\subsection{Weak Supervision}
Given the supervised fine-tuning (SFT) performance on our low-prevalence deployment tabs.~\ref{tab:WeakSFTtiny},~\ref{tab:weakSFTbase} we seek to improve it by applying our method from section~\ref{weaklysupervisedMethod}.

\begin{figure}[h!]
    \centering
    \begin{minipage}[b]{0.3\linewidth}
        \centering
        \includegraphics[width=\linewidth]{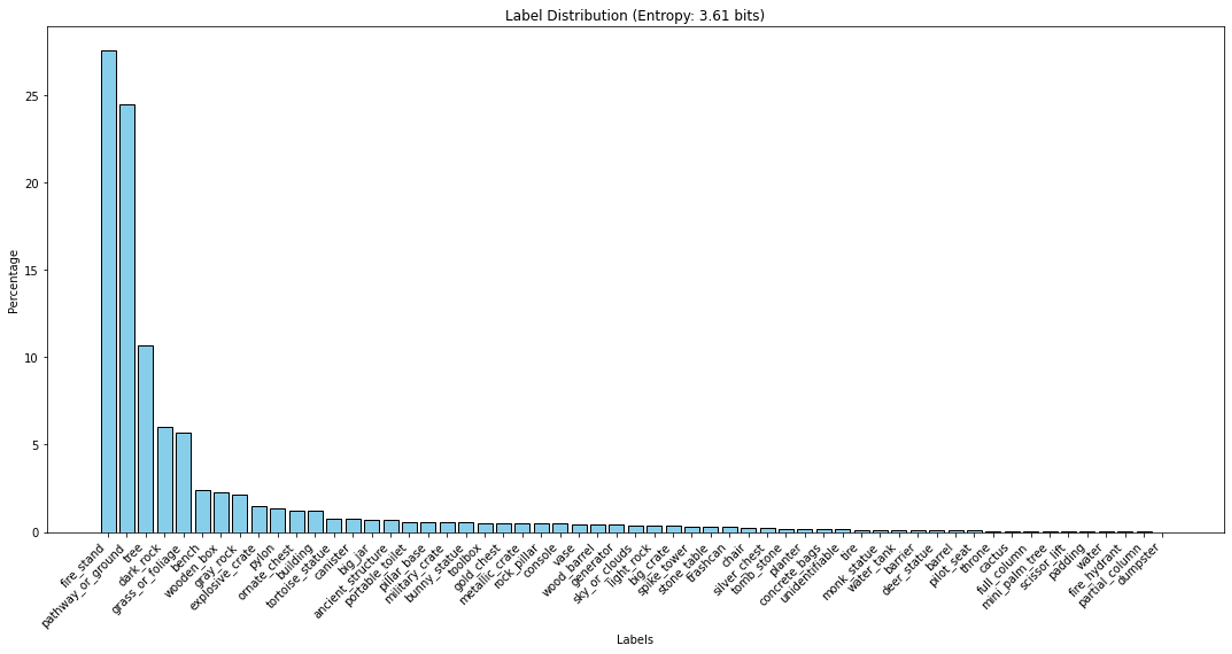}
        \textbf{(a)}
        \label{fig:hista}
    \end{minipage}
    \begin{minipage}[b]{0.3\linewidth}
        \centering
        \includegraphics[width=\linewidth]{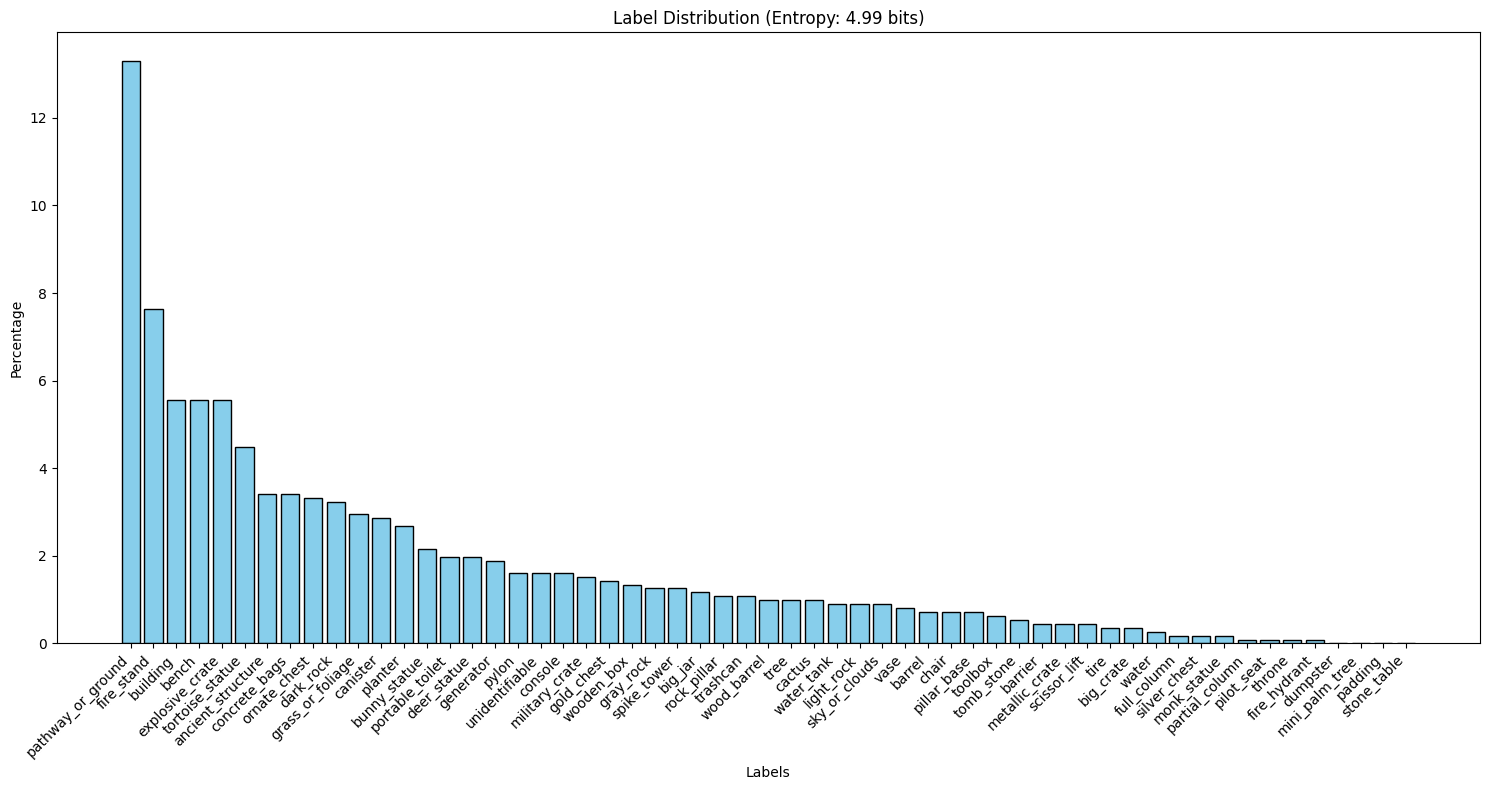}
        \textbf{(b)}
         \label{fig:histb}
    \end{minipage}

    
    \begin{minipage}[b]{0.3\linewidth}
        \centering
        \includegraphics[width=\linewidth]{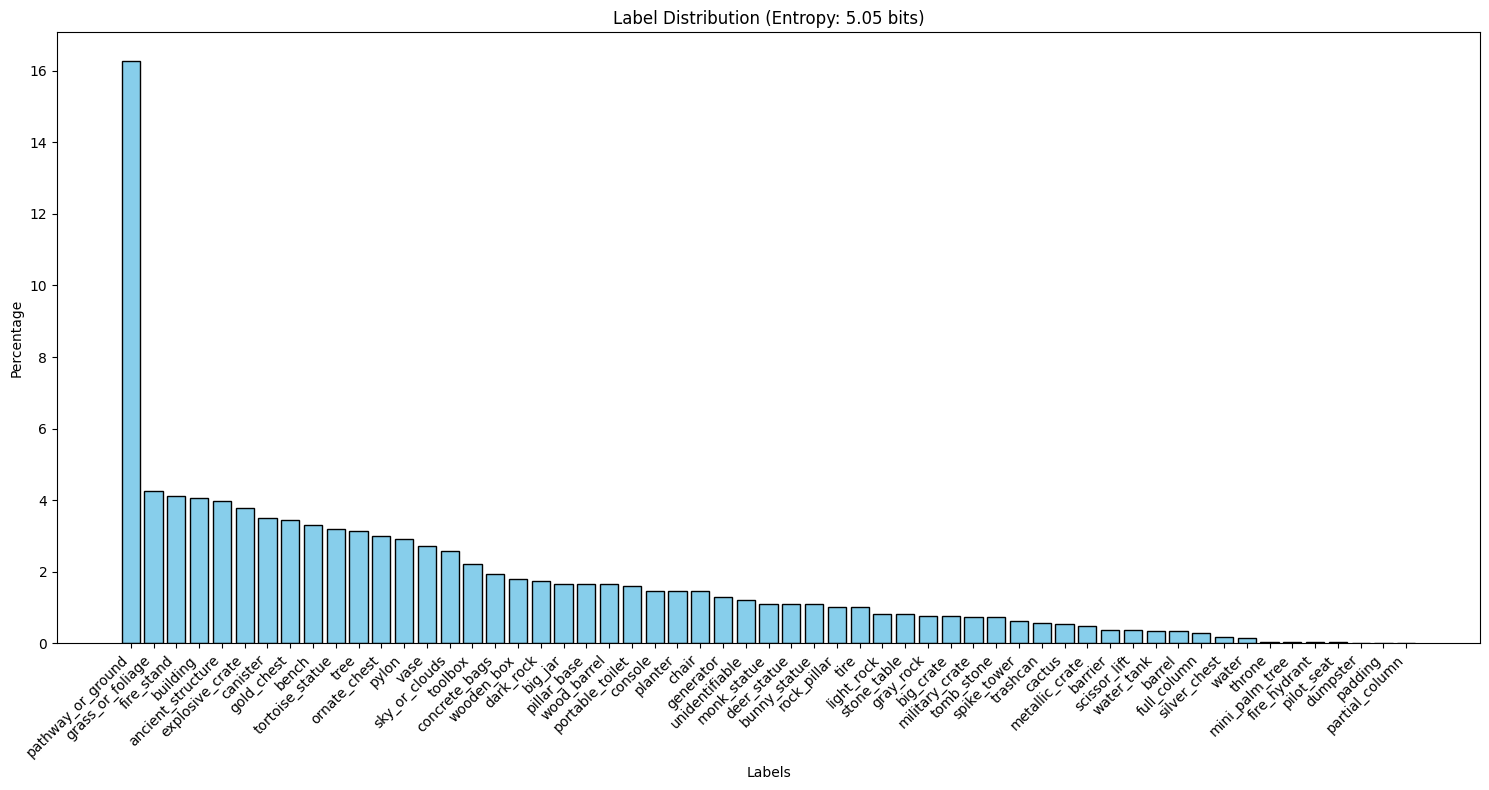}
        \textbf{(c)}
     \label{fig:histc}
    \end{minipage}
    \begin{minipage}[b]{0.3\linewidth}
        \centering
        \includegraphics[width=\linewidth]{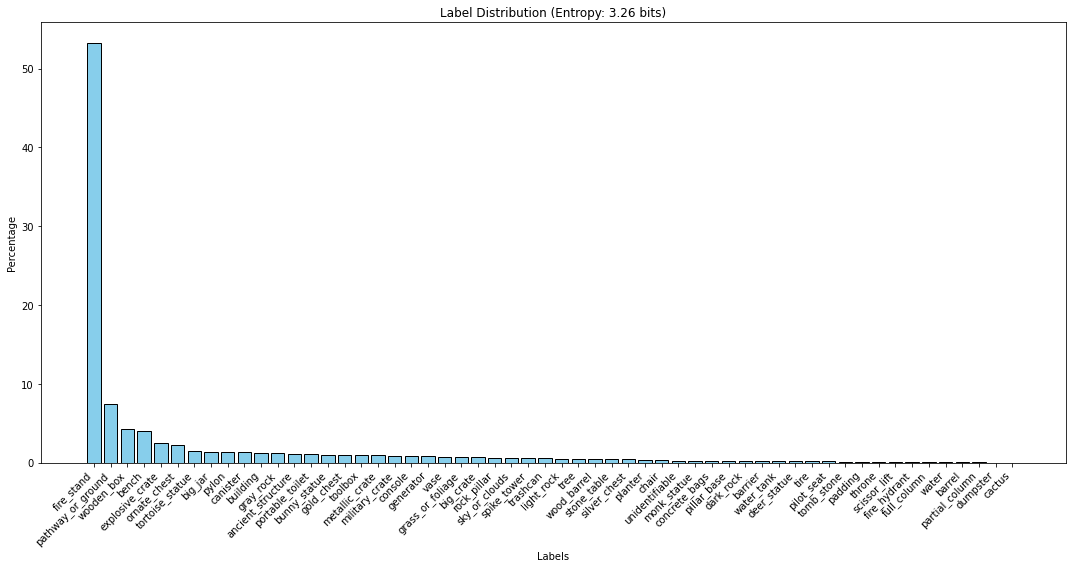}
        \textbf{(d)}
         \label{fig:histd}
    \end{minipage}
    \caption{Mask label frequencies. \textbf{(a)} ground truth \textbf{(b)} 50 re-sampled from k=50 clusters, \textbf{(c)} 50 samples re-sampled from k=100 clusters, \textbf{(d)} text prompt based filtering on semantic categories trees, foliage, roads, sky}
    \label{fig:histograms}
\end{figure}

\begin{figure}
    \centering
     \scalebox{0.4}{%
    \begin{minipage}{.45\linewidth}
        \centering
        \includegraphics[width=\linewidth]{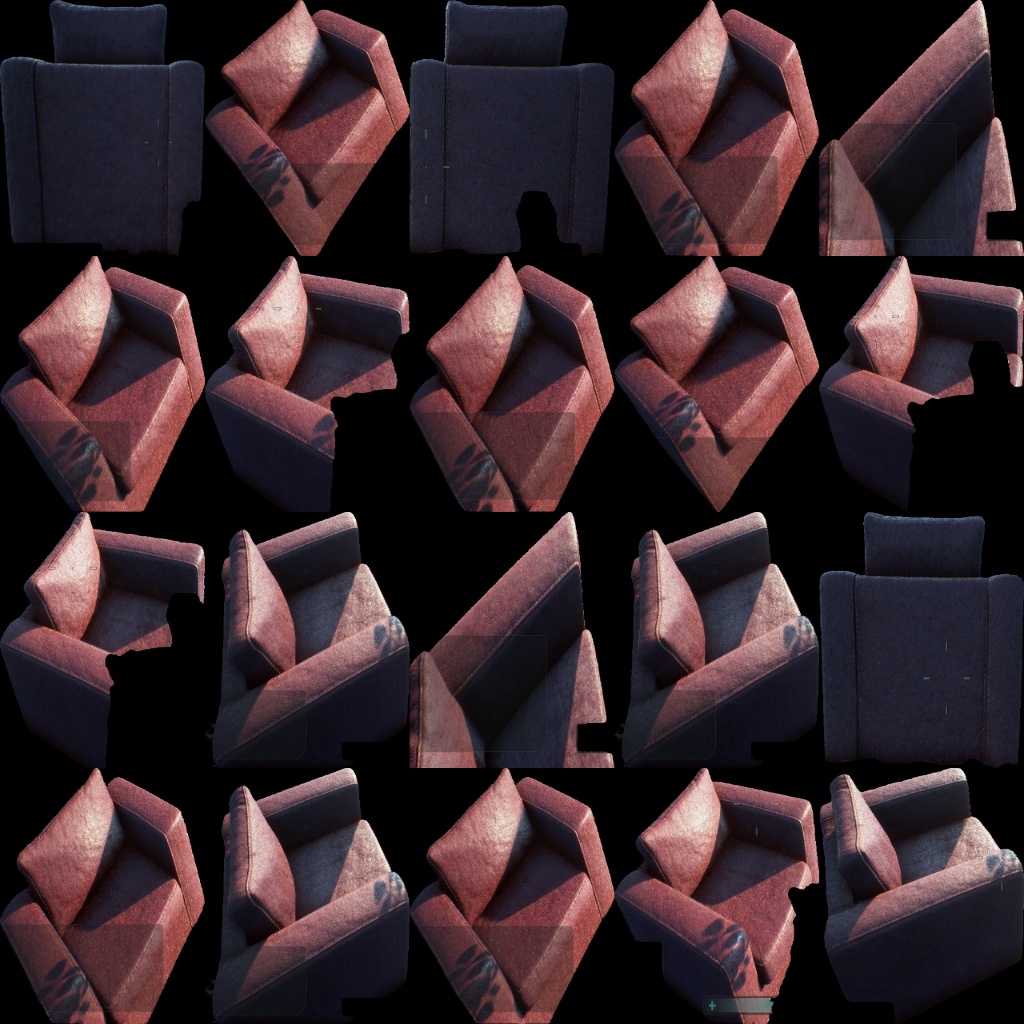}
        \textbf{(a)}
    \end{minipage}
    \hfill
    \begin{minipage}{.45\linewidth}
        \centering
        \includegraphics[width=\linewidth]{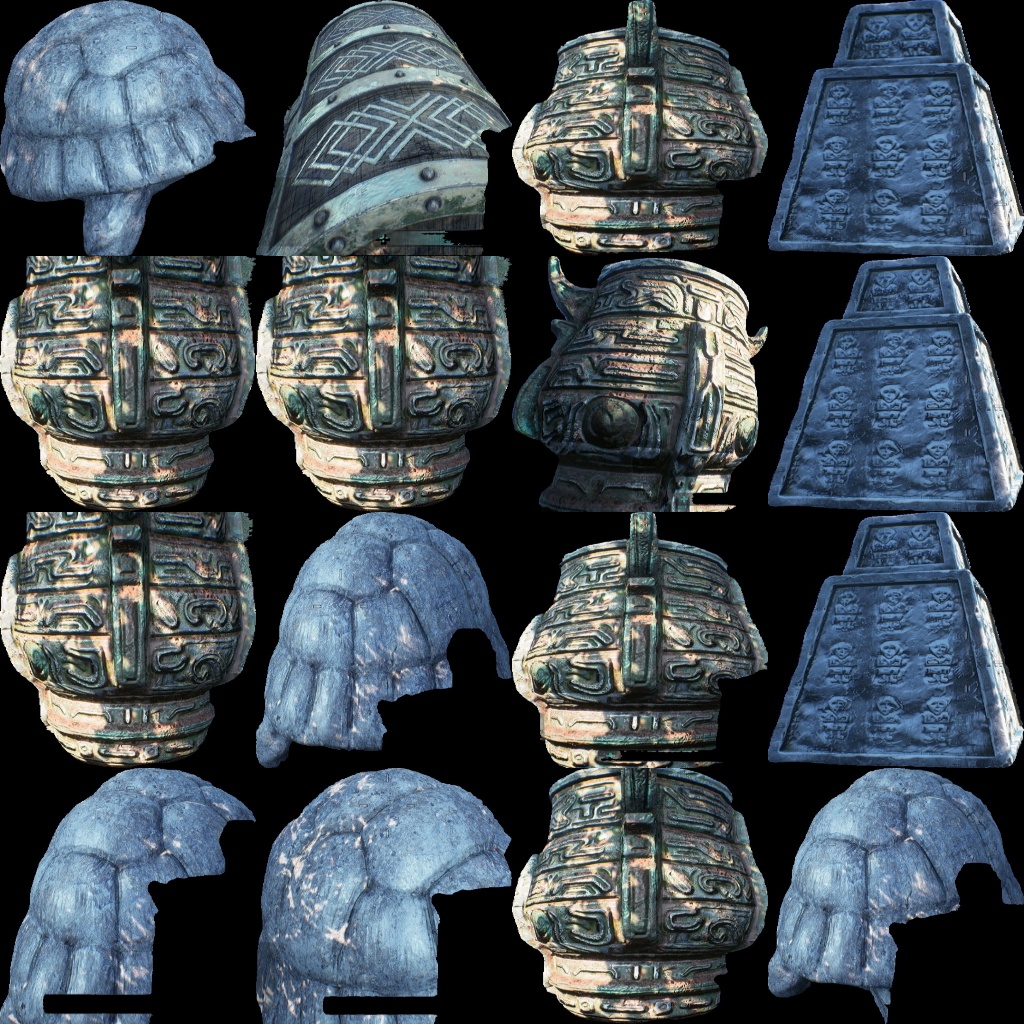}
        \textbf{(b)}
    \end{minipage}
    \begin{minipage}{.45\linewidth}
        \centering
        \includegraphics[width=\linewidth]{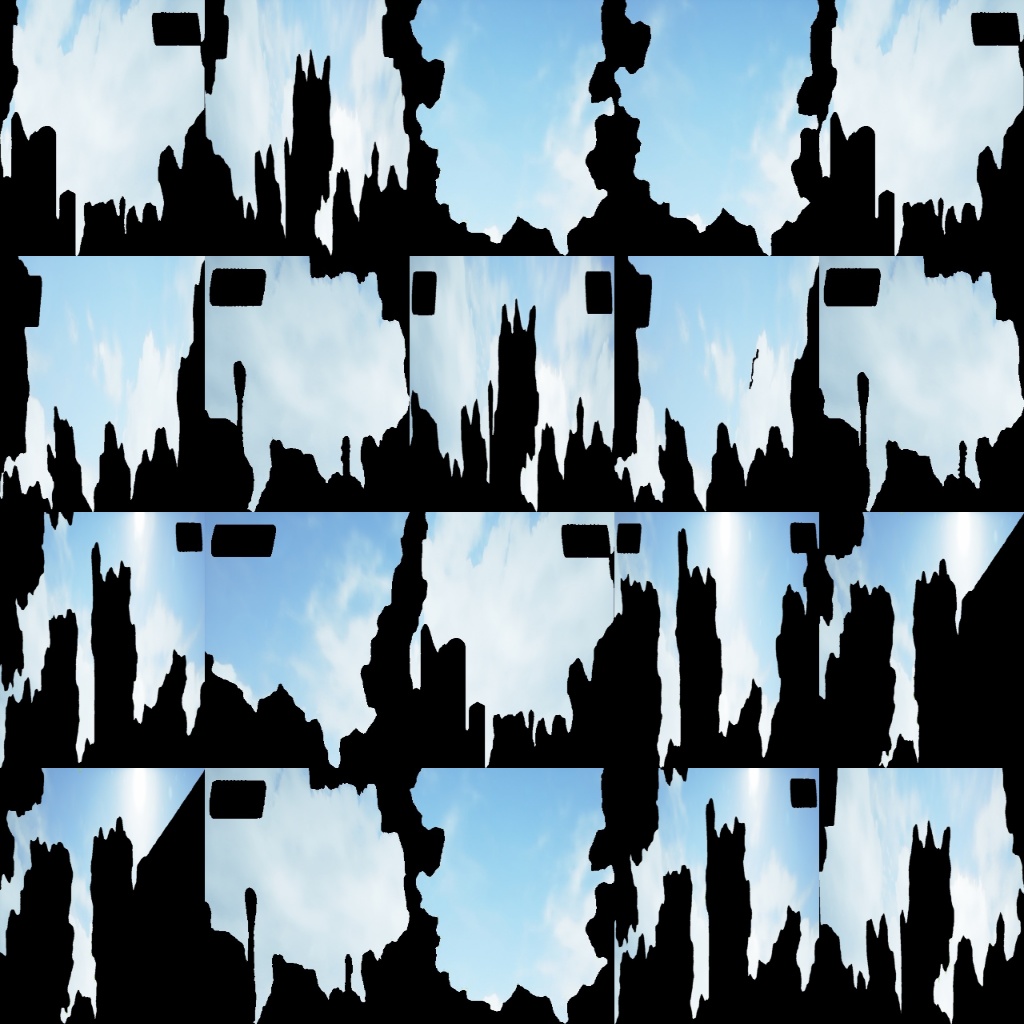}
        \textbf{(c)}
    \end{minipage}
    \hfill
    \begin{minipage}{.45\linewidth}
        \centering
        \includegraphics[width=\linewidth]{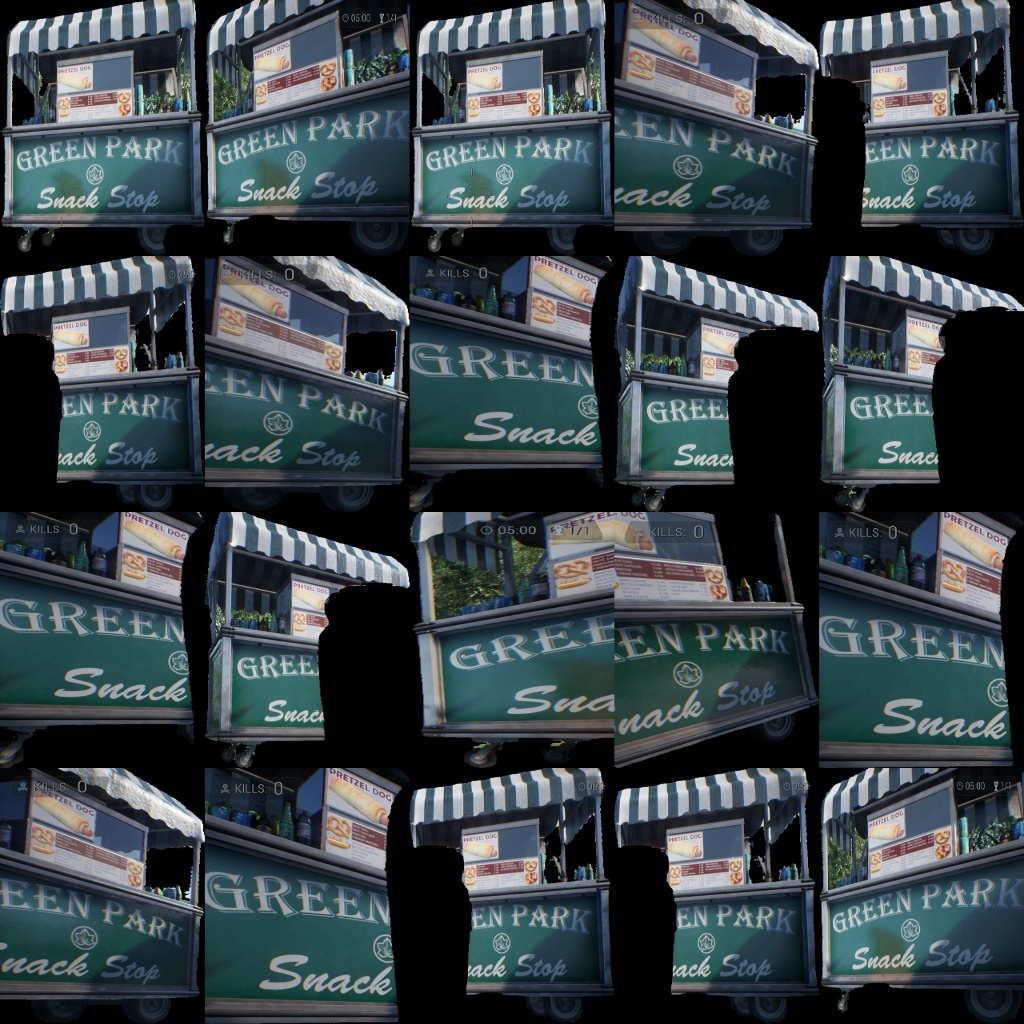}
        \textbf{(d)}
    \end{minipage}
    }
    \caption{Mask clustering(k=50): \textbf{(a)} multiple views of objects are captured, \textbf{(b)} certain fine-grained objects tend to cluster together, \textbf{(c)} the sky, an "object" not relevant to our visual bug, \textbf{(d)} map object.}
    \label{fig:ClusterQualitative}
\end{figure}

\textbf{Mask Filtering:}
 To analyze the masks produced by SAM \cite{kirillov2023segment}, we sample 30k frames from an unlabeled human gameplay video from GM50, generate masks using SAM and label them. Our labeling scheme was a combination of GM50 Objects of Interest (OOI) along with other general semantic categories. As observed in fig.~\ref{fig:histograms}a, firestand, pathway, ground, and trees dominate the distribution. The latter two are omnipresent in scenes and the former, due to the data gathering policy. This creates redundancy in the signal we inject via augmentation. To combat this, we use CLIP \cite{radford2021learningCLIP} to extract embeddings and HAC \cite{Johnson1967HierarchicalCS} ($k=50$) with cosine distance to cluster masks in an unsupervised manner, $k$ was selected naively with \textit{a priori} knowledge of 50 OOI on the map. Realistically $k > 50$ as other non-OOI are contribute to visual semantics of GM50. We observe that resampling after using either the heuristic fig.~\ref{fig:histograms}b to select $k$ or overclustering fig.~\ref{fig:histograms}c ($k=100$) somewhat ameliorates class imbalance. See fig.~\ref{fig:ClusterQualitative} for qualitative examples of our clusters. Interestingly, clusters capture multiple views of both OOI  fig.~\ref{fig:ClusterQualitative} and also other map objects  fig.~\ref{fig:ClusterQualitative}d, the food stand is not an OOI yet it is captured, a promising sign for OOD generalization. Further, we observe that objects with overlapping  visual semantics, especially fine-grained ones such as variants of statues fig.~\ref{fig:ClusterQualitative}b, tend to cluster together. We explore explicit removal of non-relevant yet highly frequent masks such as sky, trees, pathways, in hopes to further increase signal in our weak dataset. As we are already using CLIP image-encoder to extract visual features, we can pair with text encoder embeddings that may be supplied interactively or stored as a priori knowledge. eg. Clipping with grass and foliage is near universally a non-issue. We filter via stored text prompts tailored towards pathways, trees, etc. resulting in a distribution fig.~\ref{fig:histograms}d. While the non-relevant masks have been filtered, overall class balance has gotten worse. By removing omnipresent non-relevant classes ( $\sim$50\% of the masks), any remaining over-represented classes (fire stand) overwhelm the distribution. We rebalance by performing clustering and resampling post text filtering. There exist other interesting approaches not explored, eg. clustering followed by interactive labeling to prune away entire clusters.

\begin{table}
    \centering
    \begin{minipage}{0.48\linewidth}
        \centering
        \scalebox{0.55}{ 
        \begin{tabularx}{1.2\linewidth}{l l r}
            \toprule
            \textbf{Dataset} & \textbf{Train Method} & \textbf{F1} \\
            \midrule
            Supervised GM4-tiny & SFT & 0.153 \\
            TinyAug + GM4-tiny & Pretrain + SFT & 0.479 \\
            LargeAug + GM4-tiny & Pretrain + SFT & 0.397 \\
            TinyAug + GM4-tiny & multi-task & 0.484 \\
            LargeAug + GM4-tiny & multi-task & 0.484 \\
            \bottomrule
        \end{tabularx}
        }
        \caption{Low-prevalence, OOD deployment F1 results on GM50. GM4-tiny training dataset (15 labeled examples). LargeAug=17k masks, TinyAug=217 masks.}
          \label{tab:WeakSFTtiny}
    \end{minipage}
    \hfill
    \begin{minipage}{0.48\linewidth}
        \centering
        \scalebox{0.55}{
        \begin{tabularx}{1\linewidth}{l r}
            \toprule
            \textbf{Dataset} & \textbf{F1} \\
            \midrule
            LargeAug-Raw & 0.054 \\
            LargeAug & 0.429 \\
            TinyAug-Raw & 0.296 \\
            TinyHeavyAug-Raw & 0.480 \\
            TinyAug & 0.529 \\
            TinyHeavyAug & 0.493 \\
            \bottomrule
        \end{tabularx}
        }
        \caption{Low-prevalence, OOD deployment F1 scores on GM50 in few-shot setting (ie. self-supervised objective only. 5 labeled negative examples, 0 positive examples). LargeAug=17k Masks, TinyAug=217 masks. Raw suffix denotes unfiltered.}
          \label{tab:TinyDataFewGoodShots}
    \end{minipage}
\end{table}

\textbf{Self-supervision: Pre-training vs multi-task:}
 Given two mask sets, Tiny (217 masks) and Large (17k masks), we create multiple datasets to serve the self-supervised objective. The first TinyAug and LargeAug consist of paired data with limited rotation augmentation of the individual masks. The second HeavyTiny and HeavyLarge consist of heavy rotations to influence diversity. We pair these objectives with labeled GM4-tiny and GM4-base in a sequential pre-training or simultaneous multi-task training setting. The multi-task objective is a weighted combination $L = \lambda L_w + (1 - \lambda) L_t$ where $L_w$ denotes our SSL objective,  $L_t$ is the target objective. We evaluate our models in the low-prevalence OOD setting on GM50 across 3 settings, each denoting some amount of "real" labeled data available during training. \textbf{1.} only a few (5) labeled \textit{"good"} exemplars and 0 positives (i.e., 0 real bugs samples) trained with weak supervision only tab.~\ref{tab:TinyDataFewGoodShots}, \textbf{2.} tiny amount of labeled data is available (15 examples total) tab.~\ref{tab:WeakSFTtiny}, and \textbf{3.} small amount of labeled data is available (156 samples total), tab.~\ref{tab:weakSFTbase}. Our results indicate that our self-supervision alone absent any positive (bugs) examples is sufficient to surpass the best fully supervised training in the low-labeled, low-prevalence regime, 0.529 vs 0.336 F1. Further fine-tuning on a small amount of labeled data tab.~\ref{tab:weakSFTbase} enhances performance to 0.550. Overall both pre-training and multi-task are competitive with one another, however pre-training edges out. In addition, we observe that pre-training was simpler to optimize, as the loss weight ($\lambda$) is a sensitive hyperparameter.  LargeAug, created from thousands of masks produces worse results overall than Tiny which has 217 masks. This is likely due to the aforementioned distribution imbalance in the masks producing information redundant samples, further exacerbated by scale. Similarly, for raw unfiltered masks, results indicate rebalancing and filtering as a progressive step; however with the right mask augmentations, sufficient diversity is introduced to make it competitive.  We make similar observations with our method on texture bugs (ref. supplemental.)

\definecolor{Gray}{gray}{0.9}

\begin{table}
\centering
\scalebox{0.75}{ 
\begin{tabularx}{0.75\linewidth}{l l l r}
\toprule
\textbf{Dataset} & \textbf{Train Method} & \textbf{F1} \\
\midrule
Supervised GM4 & SFT & 0.336 \\
LargeAug + GM4 & multi-task & 0.419 \\
TinyHeavyAug + GM4 & multi-task & 0.516 \\
TinyHeavyAug-raw + GM4 & multi-task & 0.510 \\
LargeAug + GM4 & multi-task & 0.533 \\
TinyHeavyAug + GM4 & Pretrain+SFT & 0.492 \\
TinyAug + GM4 & Pretrain+SFT & \textbf{0.550} \\
\bottomrule
\end{tabularx}
}
\caption{Low-prevalence, OOD deployment F1 results on GM-50. GM4-base training dataset (175 total labeled examples). Multi-task and pre-training on the self-supervised objective greatly increases performance over baseline 0.336 F1 score obtained from SFT. TinyAug = small mask set. Raw suffix = unfiltered.}
\label{tab:weakSFTbase}
\end{table}

\section{Discussion, Limitations and Future work}
\vspace{-0.2cm}

Our method, which utilizes weak-supervision to scale up a self-supervised objective improves performance both through multi-task and pre-training. It consistently demonstrates superior performance compared to solely using a supervised low-labeled dataset. Our self-supervision however is domain-crafted in contrast with advances in recent general, less biased approaches \cite{he2021masked, bardes2022vicreg, caron2021emerging, oquab2023dinov2}; we only make use of unlabeled data as a means to obtain representative object centric masks. Additional information exists in unsupervised videos to be captured through general self-supervised objectives, for instance we can use rebalanced masks with DINO \cite{caron2021emerging, oquab2023dinov2} to adapt the backbone. Our GM environment has shared, yet inverted objectives to PUG \cite{bordes2023pug}; \cite{bordes2023pug} use interactive Unreal environments to serve as simulators to obtain photorealistic data in a controlled manner whereas our target distributions are the simulators themselves. A limitation of our approach is reliance on the policy under which data was gathered. The integration of Reinforcement Learning agents, such as \cite{abdelfattah2023preferenceconditioned}, is an intriguing avenue for future research. Additionally, Fig.~\ref{fig:ClusterQualitative}b highlights a challenge: our filtering approach allows text-prompts to specify preference-based semantics, yet it struggles when these semantics are fine-grained or not well-represented within the embedding. Thus, the text-image model has difficulty performing in a zero-shot context. Future work might consider advanced text-image models or exploring strategies that combine text-image prompting with other learning methods. Additionally, models adapted from SAM \cite{li2023semanticsam, cheng2023tracking} can be applied during segmentation stage to enhance extraction of semantic masks.

 The rapid testing cycles and cadence of new content make traditional label-intensive learning impractical for visual bug detection. Despite game engines increasingly integrating ML capabilities, relying solely on integration isn't scalable; our work moves towards techniques not reliant on source code access. Further, new game content can be viewed as OOD data and we have taken steps towards methods that are robust and generalize to such scenarios, specifically objects. Future work may explore the scalability and generality of our methodology across various visual bug-types and OOD settings. What data requirements exist for domain adaptation to art styles (eg. non-photorealistic games), environments, lighting? Moreover, constraining ourselves to RGB-only for practical reasons fails to exploit the richness of multimodality, limiting the depth of visual cues our models may capture. Multimodal data can be used during training and constrained or estimated at test time, maintaining practicality. Further, our augmentation strategy uses traditional CV techniques, however other synthetic or generative methods may also be an interesting line of future work.

\vspace{-0.5cm}
\section{Conclusion}
\vspace{-0.2cm}
  Visual bug detection poses unique challenges due to rapidly evolving content, constraints in labeled data availability, and generalization to out-of-distribution scenarios. In this study, we explored a weakly-supervised, three-staged approach to address these challenges, specifically targeting first-person player clipping (FPPC) within Giantmap. Our findings harness the potential of large-pretrained visual models to enhance our training data. Our approach allows for the injection of priors through prompting, both geometric and text-based. A significant advantage of promptable filtering is its simplicity, making it accessible for non-ML professionals, allowing them to integrate their expert knowledge into the self-supervised objective. Additionally, our framework shows promise in generating expansive, curated datasets within video games, with the potential to foster both, comprehensive understanding of video game scenes and developing visual bug detection models.

\bibliography{gametest}
\end{document}